\documentclass{article}
\usepackage{ijcai25}

\usepackage{times}
\usepackage{soul}
\usepackage{url}
\usepackage[hidelinks]{hyperref}
\usepackage[utf8]{inputenc}
\usepackage[small]{caption}
\usepackage{graphicx}
\usepackage{amsmath}
\usepackage{amsthm}
\usepackage{booktabs}
\usepackage{algorithm}
\usepackage{algorithmic}
\usepackage[switch]{lineno}
\usepackage{multirow}
\usepackage{makecell}

\usepackage{amsmath,amsfonts,bm}

\def\eqref#1{equation~\ref{#1}}
\def\1{\bm{1}}

\def\vmu{{\bm{\mu}}}
\def\vtheta{{\bm{\theta}}}

\def\vr{{\bm{r}}}

\def\vz{{\bm{z}}}

\def\mA{{\bm{A}}}
\def\mB{{\bm{B}}}
\def\mC{{\bm{C}}}

\def\mU{{\bm{U}}}
\def\mV{{\bm{V}}}
\def\mW{{\bm{W}}}

\DeclareMathAlphabet{\mathsfit}{\encodingdefault}{\sfdefault}{m}{sl}
\SetMathAlphabet{\mathsfit}{bold}{\encodingdefault}{\sfdefault}{bx}{n}

\def\gA{{\mathcal{A}}}

\def\gN{{\mathcal{N}}}

\def\sR{{\mathbb{R}}}

\def\sZ{{\mathbb{Z}}}

\DeclareMathOperator*{\argmin}{arg\,min}

\title{Federated Low-Rank Adaptation for Foundation Models: A Survey}

\author{ 
Yiyuan Yang$^1$
\and
Guodong Long$^1$\and
Qinghua Lu$^{2}$\and
Liming Zhu$^2$\and
Jing Jiang$^1$\And
Chengqi Zhang$^3$\\
\affiliations
$^1$University of Technology Sydney, Australia\\
$^2$CSIRO's Data61, Australia\\
$^3$The Hong Kong Polytechnic University, China\\
\emails
Yiyuan.Yang-1@student.uts.edu.au,
\{guodong.long, jing.jiang\}@uts.edu.au,
\{Qinghua.Lu, Liming.Zhu\}@data61.csiro.au,
chengqi.zhang@polyu.edu.hk,
}
\begin{document}

\maketitle

\begin{abstract}
Effectively leveraging private datasets remains a significant challenge in developing foundation models. Federated Learning (FL) has recently emerged as a collaborative framework that enables multiple users to fine-tune these models while mitigating data privacy risks.  Meanwhile, Low-Rank Adaptation (LoRA) offers a resource-efficient alternative for fine-tuning foundation models by dramatically reducing the number of trainable parameters. This survey examines how LoRA has been integrated into federated fine-tuning for foundation models—an area we term FedLoRA—by focusing on three key challenges: distributed learning, heterogeneity, and efficiency. We further categorize existing work based on the specific methods used to address each challenge. Finally, we discuss open research questions and highlight promising directions for future investigation, outlining the next steps for advancing FedLoRA\footnote{\url{https://github.com/Lydia-yang/Awesome-Federated-LoRA}}.

\end{abstract}

\section{Introduction}
With the increasing limitations of centralized learning in handling large-scale and privacy-sensitive data, Federated Learning (FL) \cite{zhang2021survey} has emerged to collaboratively learn model across distributed clients without direct access to data. One of the primary challenges in FL lies in the computational and communication overhead \cite{almanifi2023communication}, as clients must train models locally and periodically exchange updates with the central server. This challenge becomes even more pronounced with modern foundation models, which are increasingly growing in depth and size, e.g., large language models (LLMs) \cite{minaee2024large} often reaching millions or even billions of parameters. The substantial resource demands of such models necessitate the development of efficient learning techniques in FL.

\begin{figure}
    \centering
    \includegraphics[width=\linewidth]{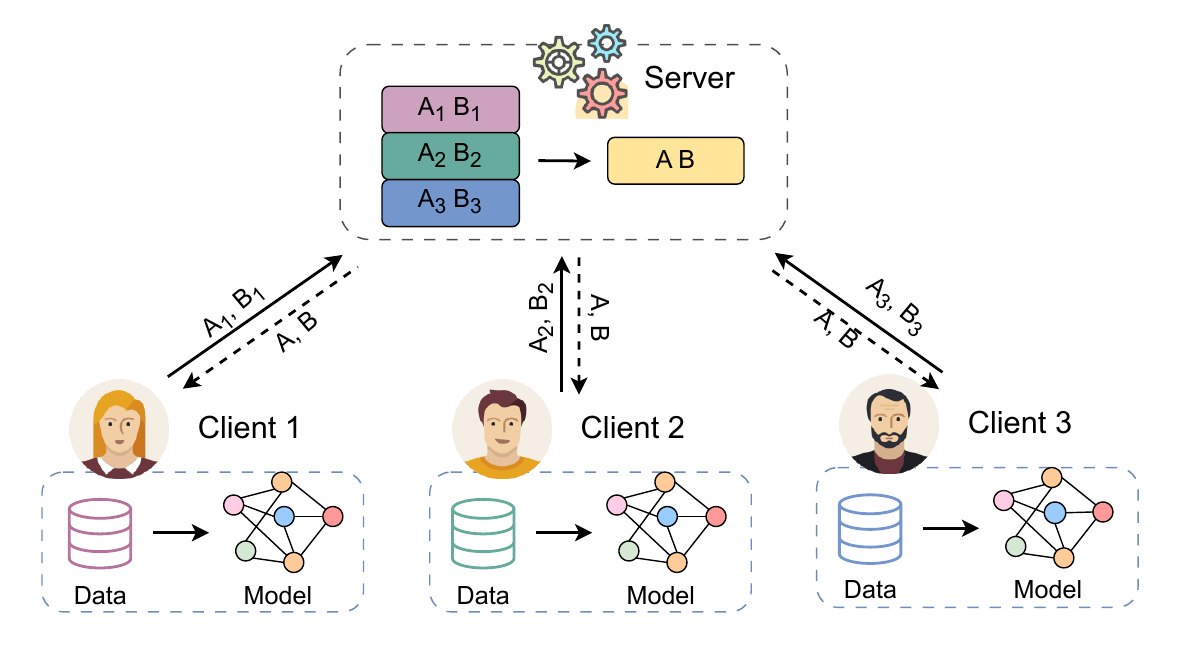}
    \vspace{-20pt}
    \caption{The overall framework of FedLoRA, where only the LoRA parameters ($A$ and $B$) are communicated for efficient learning.}
    \label{fig:fedlora-famework}
\end{figure}
Parameter-efficient fine-tuning (PEFT) methods \cite{han2024parameter} have been proposed as an effective solution for tuning foundation models, which optimize only a subset of parameters to enable efficient learning while maintaining performance comparable to full fine-tuning, thereby making it particularly suitable for resource-constrained environments. Inspired by this, recent studies \cite{kuang2024federatedscope,zhang2023fedpetuning} have explored the integration of PEFT methods into the FL framework to enhance training efficiency. Among these methods, Federated Low-Rank Adaptation (FedLoRA), integrating LoRA \cite{hu2021lora} in FL, has gained significant attention due to its efficiency and parallel ability during training processes. By decomposing model updates into low-rank matrices for learning and communicating, FedLoRA significantly reduces computation and communication costs while maintaining model performance.

Although FedLoRA addresses efficiency concerns to some extent, numerous challenges such as heterogeneity remain when aggregating LoRA in FL. In response, extensive research has been conducted to enhance the effectiveness and efficiency of FedLoRA. To systematically analyze and consolidate these advancements, we present a comprehensive survey on FedLoRA, providing an in-depth exploration of its methodologies, challenges, and solutions. Unlike existing surveys that primarily focus on broader aspects of efficient FL training \cite{almanifi2023communication,woisetschlager2024survey} or general federated foundation models \cite{zhuang2023foundation,yu2023federated,ren2024advances}, our study specifically examines the detailed mechanisms and unique challenges of LoRA aggregation in FL. By offering a focused and structured analysis, this survey aims to bridge existing research gaps and provide insights into future directions for FedLoRA.

\paragraph{Contributions.} Our contributions can be summarized as: 1) A structured
taxonomy: we present a systematic classification of FedLoRA based on different technologies; 2) A comprehensive review: we conduct an extensive survey of recent advancements in FedLoRA within the proposed taxonomy; 3) Some future directions: we highlight open challenges and emerging research opportunities in FedLoRA. 

\begin{figure}
    \centering
    \includegraphics[width=\linewidth]{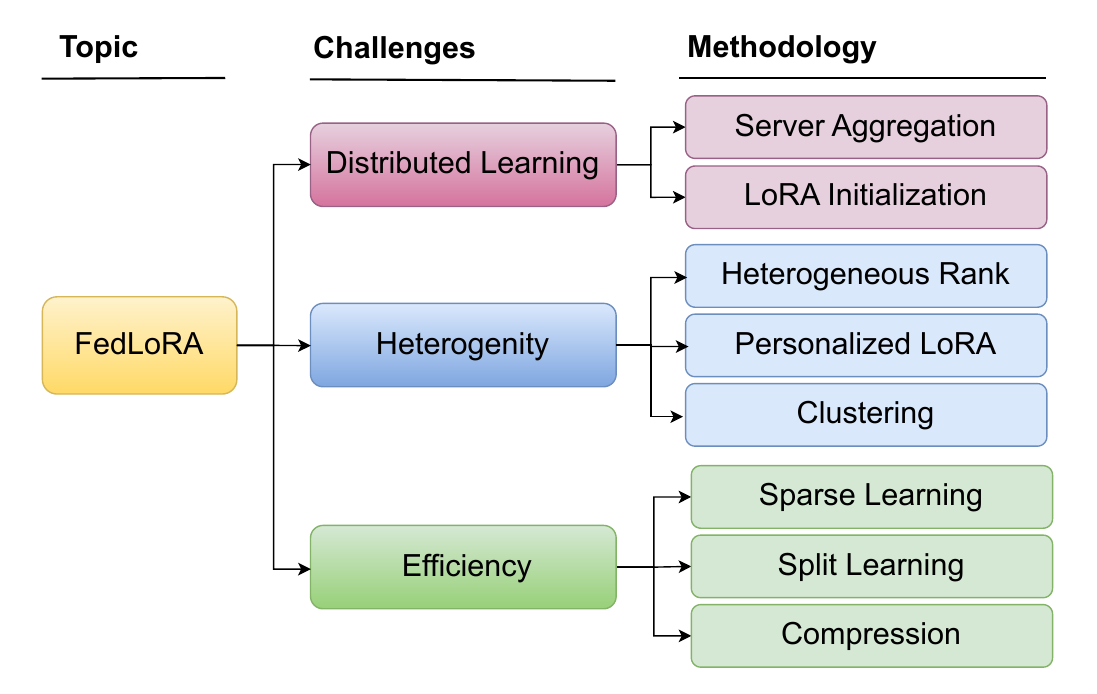}
    \vspace{-20pt}
    \caption{Taxonomy of FedLoRA focusing on distributed learning, heterogeneity and efficiency with further classified subcategories.}
    \label{fig:fedlora-taxonomy}
\end{figure}

\section{Preliminary}
\subsection{Federated Learning}
FL \cite{zhang2021survey} is a decentralized machine learning paradigm designed to enable multiple devices or clients to collaboratively train models while preserving privacy data by ensuring that data remains localized on each client. Typically, in an FL scenario, there are $K$ clients in an FL scenrio, where each client $k$ has access to its own local dataset $D_k$. The objective of FL is to optimize a global model by minimizing a weighted sum of local objective functions across all clients:
\begin{equation}
    \min_{\mW} f(\mW) = \sum_{k=1}^K p_k f_k(\mW;D_k),
\end{equation}
where $f(\mW)$ represents the global objective function parameterized by $\mW$, $f_k(\mW;D_k)$ is the local objective function computed on client $k$'s dataset $D_k$, and $p_k$ is the weight assigned to client $k$. Building on this formulation, numerous FL methods have been developed to tackle a variety of challenges inherent to the paradigm, including data heterogeneity arising from varying client data distributions, communication efficiency necessitated by frequent model updates, and so on.

\subsection{Low-Rank Adaptation}
\label{sec-LoRA}
With the rapid growth in the size and depth of modern foundation models, PEFT methods \cite{han2024parameter} have emerged as a practical solution to adapt these large-scale models efficiently by learning only a small subset of parameters. Among these PEFT methods, LoRA \cite{hu2021lora} distinguishes itself through its inherent parallelism and efficiency, achieving impressive results without introducing new parameters into the model. LoRA leverages low-rank decomposition by reparameterizing the weight updates for each layer, thereby significantly reducing memory consumption and computational overhead while maintaining model performance. Formally, given a weight matrix $\mW \in \sR^{m \times n}$, LoRA introduces two low-rank matrices, $\mA \in \sR^{r \times n}$ and $\mB \in \sR^{m \times r}$, where $r \ll \min(m,n)$, which can be formulated as:
\begin{equation}
    \mW = \mW{'} + \Delta\mW = \mW{'} + \mB\mA.
\end{equation}
By restricting the rank $r$, LoRA ensures that the number of additional parameters is significantly smaller than the original model size, enabling efficient fine-tuning of large-scale models without sacrificing performance.% Its effectiveness has been widely demonstrated across diverse tasks and domains.

\label{loras}
\paragraph{Variations of LoRA.} Various adaptations of LoRA have been proposed to address different challenges and applications, including dynamic rank adjustment for adaptability \cite{valipour2022dylora,zhang2023adalora}, quantization techniques for further efficiency \cite{dettmers2024qlora,xu2023qa}, and advanced strategies such as kernel-wise adaptation \cite{chen2022empowering} and Bayesian inference \cite{yang2023bayesian} to enhance overall effectiveness.
% There are numerous variations of LoRA tailored to address different challenges and application scenarios. For example, some previous studies 
% \cite{valipour2022dylora,zhang2023adalora,zhang2023increlora}
% \cite{valipour2022dylora,zhang2023adalora}
% have proposed methods to dynamically adjust the rank of LoRA to enhance its adaptability to diverse tasks and data distributions. Additionally, a series of works %\cite{dettmers2024qlora,xu2023qa,li2023loftq}
% \cite{dettmers2024qlora,xu2023qa}
% have integrated quantization techniques to further improve the efficiency of LoRA. Furthermore, several innovative techniques, such as kernel-wise adaptation \cite{chen2022empowering} and Bayesian inference \cite{yang2023bayesian}, have been incorporated into LoRA to further enhance its effectiveness.

\subsection{Federated Low-Rank Adaptation}
With the increasing demand for efficient learning in FL, recent research has started to explore the adaptation of LoRA within FL (FedLoRA). FedLoRA leverages the core idea of LoRA by learning and communicating only a small subset of parameters, thereby enhancing both computational and communication efficiency. The overall architecture can be seen in Figure~\ref{fig:fedlora-famework}, where each client maintains its own model for local training while only the LoRA parameters are transmitted to the server for aggregation. The overall objective is:
\begin{equation}
    \min_{\Delta \mW} f(\mW) = \sum_{k=1}^K p_k f_k(\mW',\Delta\mW;D_k),
\end{equation}
where $\Delta\mW=\mB\mA$ denotes the learnable LoRA parameters.% of two low-rank matrices.
% While numerous surveys \cite{} have comprehensively reviewed efficient learning techniques in FL, they often overlook the potential new challenges introduced by FedLoRA. To bridge this gap, we present a comprehensive survey that systematically examines the principles, challenges, and advancements of FedLoRA, providing valuable insights for researchers and practitioners interested in this emerging field.

\paragraph{Taxonomy.}  Unlike previous surveys \cite{almanifi2023communication,woisetschlager2024survey} that provide a broad overview of efficient FL, our paper focuses specifically on the adaptation of LoRA in FL and delves into the unique challenges introduced by FedLoRA. We propose a taxonomy in Figure~\ref{fig:fedlora-taxonomy} that categorizes these challenges into three main aspects: 1) enabling effective distributed learning for stable convergence and knowledge sharing, 2) addressing heterogeneity arising from non-IID data distributions among clients to ensure stable performance, and 3) further optimizing computational and communication efficiency to improve resource utilization. Various approaches have been proposed to address these, which are further classified into subcategories based on the underlying technologies used, providing a detailed examination of the current landscape. 

\section{Distributed Learning}
As a distributed learning paradigm, FL relies on effective client aggregation and proper initialization—both before training and at each communication round—to ensure model convergence and performance consistency. This section systematically examines FedLoRA from these perspectives. %analyzing various aggregation strategies and initialization techniques to address the unique challenges posed by LoRA in FL.
% Heterogeneity is one of the fundamental challenges in FL, mainly arising from non-IID data across clients, which could lead to performance degradation, slower convergence, and imbalanced contributions. This section explores recent advancements in FedLoRA that address heterogeneity through server aggregation, LoRA initialization, and personalization. %In this section, we focus on recent advancements that adapt LoRA to the FL paradigm (FedLoRA) by proposing novel approaches to address the heterogeneity, which can be broadly categorized into three key areas: server aggregation, LoRA initialization and personalization.

\subsection{Server Aggregation}
\label{sec-aggre}
To facilitate knowledge sharing in distributed learning systems, conventional FL often employs specific weighting algorithms, such as FedAVG \cite{mcmahan2017communication}, for server aggregation to balance contributions from diverse clients and improve global model convergence. Recent research \cite{zhang2024towards} has extended these algorithms to FedLoRA for distributed learning. However, due to the unique structure of LoRA, directly applying FedAVG in FedLoRA can result in suboptimal aggregation outcomes \cite{sun2024improving}.

\paragraph{LoRA Aggregation Discordance.}
As introduced in Section~\ref{sec-LoRA}, LoRA contains two low-rank matrices $\mA$ and $\mB$ for learning and communicating. However, applying weighting algorithms separately to matrices $\mA$ and $\mB$ is inconsistent with the objective of joint optimization, potentially leading to performance degradation. To illustrate this, consider a scenario with two clients, each with LoRA parameters $(\mA_1,\mB_1)$ and $(\mA_2,\mB_2)$ for aggregation, this discordance between separate aggregation of $\mA$ and $\mB$ and their intended joint optimization can be mathematically formulated as:
\begin{equation}
\label{lora-aggregation}
\begin{aligned}
    \underbrace{\mW = \mW'+\frac{1}{2}(\mB_1+\mB_2)\times\frac{1}{2}(\mA_1+\mA_2)}_{\text{Separate aggregation of $\mA$ and $\mB$}} \\
    \neq \underbrace{\mW'+\frac{1}{2}(\mB_1\mA_1+\mB_2\mA_2) = \mW^{*}.}_{\text{Ideal aggregation for joint optimization}}
\end{aligned}
\end{equation} 
\paragraph{Single Low-Rank Matrix Aggregation.}
One simple and intuitive approach to addressing this LoRA aggregation discordance is to learn and aggregate only one low-rank matrix per communication round. Specifically, during each communication round, only one low-rank matrix (either $\mA$ or $\mB$) is learned and sent to the server for aggregation, while the other matrix remains fixed and consistent across all clients.
This strategy ensures that the ideal aggregation can be equivalently achieved by aggregating a single low-rank matrix, thereby simplifying the process and maintaining computational efficiency. It can be formulated as follows:
\begin{equation}
\label{low-rank-aggregation}
\begin{aligned}
    \mW'+\frac{1}{2}(\mB\mA_1+\mB\mA_2) = \mW'+\frac{1}{2}\mB(\mA_1+\mA_2) \\
    \text{or }
    \mW'+\frac{1}{2}(\mB_1\mA+\mB_2\mA)  =\mW'+\frac{1}{2}(\mB_1+\mB_2)\mA.
\end{aligned}
\end{equation} 
The study \cite{sun2024improving} first analyzed the discordance issue in FedLoRA and proposed FFA-LoRA, which freezes the low-rank matrix $\mA$ and only updates the matrix $\mB$ for aggregation. This approach not only achieves more consistent performance by aligning with the objective of joint optimization but also significantly reduces computational costs. Similarly, CoLR \cite{nguyen2024towards} introduced a novel strategy where, in each communication round, clients learn a newly initialized matrix $\mA$ from their local data for aggregation, while keeping $\mB$ unified across clients via decomposing full matrix $W$ on the server. To further enhance the performance of FedLoRA, RoLoRA \cite{chen2024robust} employed an alternating minimization approach, learning and aggregating only $\mB$ in odd communication rounds and $\mA$ in even rounds. This alternating strategy effectively addresses the discordance issue while providing more robust performance across heterogeneous scenes. Meanwhile, LoRA-A$^2$ \cite{koo2024towards} also explored the alternating minimization approach and incorporated an adaptive rank selection strategy to further reduce communication costs by dynamically selecting the most important LoRA ranks for learning and aggregation. 

% Although these methods are intuitive and effective to some extent, they may suffer from a lower convergence rate due to the reduced number of learnable parameters, which can limit the model’s capacity in complex heterogeneous scenarios.

\paragraph{Full-size Matrix Aggregation.} Another class of approaches to address the discordance issue is leveraging the full-size matrix $\Delta \mW$, reconstructed from the product of two low-rank matrices $\mA$ and $\mB$. Instead of aggregating $\mA$ and $\mB$ separately, this approach aggregates the full-size matrix $\Delta \mW=\mB\mA$, achieving ideal aggregation directly:
\begin{equation}
\label{full-aggregation}
\begin{aligned}
    \mW' + \Delta \overline{\mW}=\mW' + \frac{1}{2}(\Delta \mW_1 +\Delta \mW_2 )\\
    = \mW'+\frac{1}{2}(\mB_1\mA_1+\mB_2\mA_2),
\end{aligned}
\end{equation} 
where $\Delta \overline{\mW}$ represents the newly aggregated full-size matirx, and the updated low-rank matrices $\mA$ and $\mB$ can be obtained by decomposing$\Delta \overline{\mW}$.
FlexLoRA \cite{bai2024federated} was the first to adopt this approach, simultaneously learning both $\mA$ and $\mB$, aggregating the full-size LoRA weight $\Delta \mW = \mB\mA$ based on individual client contributions, and then employing Singular Value Decomposition (SVD) for weight redistribution. This method not only addressed the discordance issue but also the heterogeneous rank configurations across clients to enhance aggregation effectiveness. Similarly, FedPipe \cite{fang2024automated} also explored full-size LoRA weight aggregation and integrated quantization techniques to improve training efficiency by quantizing local models into different bit levels. At the same time, FloRA \cite{wang2024flora} implemented the full-size LoRA weight aggregation in another way, which stacks all clients' $\mA$ and $\mB$ respectively to derive the final aggregated full-size LoRA weight. 

% These methods, which leverage full-size parameters, can ensure a stable overall convergence rate. However, they would require higher computational and memory resources due to the increased cost of matrix multiplications and the decomposition of all LoRA parameters, making them less efficient for large-scale FL environments.

\paragraph{Corrective Mechanism.}
To preserve the advantages of initialization with averaged low-rank matrices \cite{bian2024lora}, a corrective mechanism is proposed to address the inaccuracies in separately aggregated LoRA matrices, bringing them closer to the ideal aggregation. 
% This approach retains the separately aggregated low-rank matrices to provide better initialization for the next communication round while correcting errors in the current aggregation process. 
FedEx-LoRA \cite{singhal2024exact} first introduced a corrective mechanism by calculating the residual $\Delta \hat{\mW}$ between the ideal aggregated weight and the weight obtained from separate aggregation, and then adding the residual to the pre-trained weight matrix $\mW'$ to rectify inaccuracies in the aggregation process, improving the alignment with the ideal aggregation. It can be formulated as:
\begin{equation}
\label{corrective-aggregation}
\begin{aligned}
    \Delta \hat{\mW} & = (\mB_1\mA_1+\mB_2\mA_2)-(\mB_1+\mB_2)(\mA_1+\mA_2), \\
    \mW & = \mW' +\Delta \hat{\mW} + \mB\mA,
\end{aligned}
\end{equation} 
where $\Delta \hat{\mW}$ represents the residual error, and $\mA$ and $\mB$ are the separately aggregated LoRA matrices. 
%the corrective mechanism aims to minimize this residual error to achieve an aggregation outcome closer to the ideal.
Additionally, LoRA-FAIR \cite{bian2024lora} also adapted the corrective mechanism by incorporating a corrective term $\Delta \mB$ to refine the matrix $\mB$, which is to minimize the residual error $\Delta \hat{\mW}$ by optimizing the similarity between the ideal aggregated LoRA weights $\Delta \overline{\mW}$ and the corrected LoRA weights $(\mB+\Delta\mB)\mA$.

% These methods preserve the initialization advantages of separately aggregated approaches while also ensuring corrective aggregation of LoRA. However, they come at the cost of increased communication overhead and additional parameters in the learning process, which may lead to higher computational complexity and resource consumption in FL settings.

\paragraph{Rank Clustering Aggregation.}
Rather than treating the low-rank matrix as a single aggregation unit, FedInc \cite{qin2024fedinc}  proposed considering each rank of LoRA as the smallest semantic unit and introduced a clustering-based aggregation algorithm. 
This approach combines the two low-rank matrices $\mA$ and $\mB$ based on their ranks to determine the newly aggregated LoRA, enabling more fine-grained and adaptive aggregation.
Specifically, $\mA \in \sR^{r\times n}$ and $\mB \in \sR^{m\times r}$ are merged into $\mC \in \sR^{r\times(m+n)}$, which represents the combined low-rank space. From $K$ clients, a set of vectors $\{\vz_i\}^{K\times r}_{i=1}$ is collected and clustered into $N$ clusters, resulting in new rank $r=N$ for the aggregated LoRA, formulated as: 
\begin{equation}
\label{cluster-aggregation}
\begin{aligned}
    \{\mC_k = \begin{bmatrix} \mA_k, \mB_k^T \end{bmatrix}\}_{k=1}^K \to \{\vz_i\}^{K\times r}_{i=1},\\
    \min_{\vmu} \sum_{n=1}^N\sum_{\vz \in \sZ_n}||\vz-\vmu_n||^2,
\end{aligned}
\end{equation} 
where $\sZ$ denotes a cluster of vectors from $\{\vz_i\}^{K\times r}_{i=1}$, and $\vmu_n$ is the centroid of cluster $\sZ_n$. The final set of centroids $\{\vmu_n\}_{n=1}^N$is used as the aggregated parameters and can be further decomposed into the newly updated matrices $\mA$ and $\mB$ with a new rank $r=N$. This method not only effectively addresses the discordance issue but also allows the LoRA rank to adapt dynamically to the inherent heterogeneity in FL.
% However, it introduces additional hyperparameters related to cluster formation and rank selection, which may require careful tuning to balance computational efficiency and model performance.
% \paragraph{Discussion.} 

\subsection{LoRA Initialization}
\label{sec-initial}
% FedLoRA utilizes LoRA as a PEFT method for efficient learning in FL,
In FL, initialization plays a crucial role in determining training efficiency, convergence speed, and model performance, and it can be broadly categorized into server-side and client-side initialization.
% In FL, the initialization of LoRA can be divided into two aspects: server-side initialization and client-side initialization. 
Server-side initialization focuses on the initial setup of LoRA parameters before training begins to ensure consistency across clients, while client-side initialization deals with the reinitialization of LoRA for individual clients at the beginning of each communication round, balancing global consistency and local adaptability. We detail these initialization strategies of FedLoRA in this section.
% The following sections provide a detailed introduction to both server-side and client-side initialization strategies for FedLoRA.

\paragraph{Server-Side Initialization.} For server-side initialization, FedLoRA typically uses a random Gaussian initialization for $\mA$ and zero initialization for $\mB$, as formulated below: 
\begin{equation}
\label{lora-init}
\begin{aligned}
    \mA = \gN(0,\sigma^2)\quad \text{and} \quad \mB = 0,
\end{aligned}
\end{equation} 
where $\gN$ denotes a Gaussian distribution with mean $0$ and variance $\sigma^2$. However, in the distributed FL setting, this standard server-side initialization can lead to significant weight update drift, widening the performance gap compared to fully fine-tuned models.
To address this challenge, SLoRA \cite{babakniya2023slora} introduced a novel data-driven initialization technique, which begins with sparse fine-tuning to find a mature starting point for LoRA, and then applies SVD to the fine-tuned parameters to initialize the low-rank matrices $\mA$ and $\mB$ for continual conventional LoRA tuning in FL. Similarly, FeDeRA \cite{yan2024federa} addressed this challenge by initializing these matrices directly from the SVD of the pre-trained weight matrices, effectively mitigating weight divergence. Overall, these methods can be generalized into:% the following framework:
\begin{equation}
\label{pretrain-init}
\begin{aligned}
    W_{pre} & \stackrel{SVD}{\longrightarrow}  \mU\Sigma\mV^T,\\
    \mA = \mV[:,:r] & \quad \text{and} \quad \mB = \mU[:r,:]\Sigma[:r],
\end{aligned}
\end{equation} 
where $W_{pre}$ is the pre-trained weight matrix for initialization. %This approach leverages the compact representation provided by SVD to produce low-rank matrices tailored to the pre-trained model, ensuring better alignment with heterogeneous client data and facilitating more efficient training.

\paragraph{Client-Side Initialization.} For client-side initialization, LoRA-FAIR \cite{bian2024lora} investigated three strategies:% for FedLoRA: Avg-Initial, Re-Initial and Local-Initial. 
\begin{itemize}
\item Avg-Initial follows conventional FL approaches, initializing clients with aggregated LoRA parameters for next communication round. For each communication round $t$ with $K$ clients, this can be formulated as $\mA_t^k=\mA_t = \sum_{k=1}^Kp_k \mA_{t-1}^k$ and $ \mB_t^k=\mB_t = \sum_{k=1}^Kp_k \mB_{t-1}^k$. %where $p_k$ represents the importance weight of each client $k$. 

\item Re-Initial reinitializes the LoRA parameters for each client $k$ in every communication round $t$ by using the formulation in Equation~\ref{lora-init} and updates each client's local pre-trained matrix by adding the aggregated LoRA parameters, denoted as $\mW'_{t,k}=\mW'_{t-1,k}+\mB_t\mA_t$, where $\mA_t$ and $\mB_t$ are aggregated matrices from all clients.

\item Local-Initial randomly selects one client $c$'s local LoRA parameters from the previous round $t-1$ as the initialization for all clients in the next round $t$, represented as $\mA_t^k=\mA_{t-1}^c, \mB_t^k=\mB_{t-1}^c, \forall k \in [K]$.

\end{itemize}
Experiments demonstrated that Avg-Initial yields the best performance due to its ability to balance continuity and unification across clients for reducing initialization drift. Building on this insight, LoRA-FAIR maintained separate aggregation for $\mA$ and $\mB$ while learning a corrective term $\Delta \mB$ to address the discordance introduced by separate aggregation.

Addtionaly, FedLoRU \cite{park2024communication} introduced a momentum-based initialization approach, extending the Re-Initial strategy from every round to periodic every $\tau$ rounds. Specifically, after each $\tau$ rounds, the server aggregates low-rank updates from clients to compute the global update $\mB\mA$, accumulates this global update with the pre-trained weight $\mW'$, and reinitializes $\mA$ and $\mB$ for the next cycle. The global model at communication round $T$ can be expressed as:
\begin{equation}
\label{mome-init}
\begin{aligned}
    \mW_T = \mW'+ {\sum_{\substack{t=1\\
    t \;\text{mod}\;\tau=0}} ^T}\mB_t\mA_t.
\end{aligned}
\end{equation} 
This approach constrains client-side optimization to a low-rank subspace by reinitializing LoRA parameters every $\tau$ communication rounds, and tailors the global model to a higher-rank space by accumulating updates from previous rounds, effectively balancing the trade-off between local adaptability and global consistency. %thereby improving performance in heterogeneous FL environments.
% By analyzing the rank properties of the loss landscape in FL, FedLoRU constrained client-side optimization to a low-rank subspace by LoRA as an implicit regularization and employed momentum-based initialization to accumulate LoRA updates from clients to form a higher-rank global model for better performance. 
% \paragraph{Discussion.} 

\subsection{Discussion} 
Effective LoRA adaptation in FL relies on advanced aggregation and initialization, and there are already numerous studies addressing these challenges as summarized in Table~\ref{tab:hetero}. Despite their effectiveness, gaps remain in the determination of importance weighting for aggregation and the theoretical understanding of its impact on model convergence and stability, which prompts future research to explore more adaptive aggregation mechanisms and establish theoretical guarantees for robustness and efficiency in FedLoRA.
%Conventional FL aggregation methods like FedAVG introduce LoRA aggregation discordance, prompting new techniques such as Single Low-Rank Matirx Aggregation for efficiency, Full-Size Matrix Aggregation for stable convergence, Corrective Mechanism for improved initialization and Rank Clustering for fine-grained aggregation (Section~\ref{sec-aggre}). Beyond these, LoRA initialization also influences model stability and performance, where server-side initialization focuses on better data alignment and client-side initialization explores the balance between local adaptability and global consistency (Section~\ref{sec-initial}). A comparative analysis of each method's advantages and disadvantages is in Table~\ref{tab:hetero}.

\section{Heterogeneity}
% \subsection{Personalization}
Heterogeneity is another key challenge in FL, mainly arising from non-IID data, which could lead to performance degradation and slower convergence. 
To address this, personalization is introduced to tailor global models to individual client's alignment. Early work \cite{yi2023fedlora} adapted LoRA as a personalization method for heterogeneous FL, which treats full-size parameters as client-specific and aggregates LoRA globally. However, as models scale, recent research has shifted toward advanced personalization techniques that focus exclusively on adapting LoRA while keeping the rest model frozen, ensuring both computational efficiency and effective learning, and we detail these methods in this section.

\subsection{Heterogeneous Rank} 
\label{sec-hetero-rank}
Considering the heterogeneous system capabilities and data distributions in FL, HETLORA \cite{cho2024heterogeneous} introduced heterogeneous LoRA ranks, allowing each client to select a personalized rank based on its task complexity and available computational resources. These heterogeneous LoRA are efficiently aggregated and distributed by local rank self-pruning and sparsity-weighted aggregation at server with objective:
\begin{equation}
\label{heter-rank}
\begin{aligned}
    \min_{\{ \mA_k,\mB_k\}} \sum^K_{k=1} p_kf_k(\mA_k,\mB_k,\mW';D_k). 
\end{aligned}
\end{equation} 
%where $f_k$ denotes the local objective function for client $k$, $D_k$ represents the dataset of client $k$, and $p_k$ is the weight assigned to client $k$. 
Building on this, subsequent research \cite{chen2024rbla,byun2024towards} enhanced it by replacing zero-padding strategy with replication-based padding strategy during aggregation, which better preserves valuable information from clients with high-quality data for better performance.
% To further enhance personalization, PF2LoRA \cite{} proposed a two-level LoRA framework consisting of one LoRA $\{\mA,\mB\}$ with a homogeneous rank for global aggregation and another $\{\mA_k,\mB_k\}$ with heterogeneous ranks for personalization. This two-level approach dynamically determines the specific rank for each client based on its training data, reducing hyperparameter complexity while achieving more robust performance than previous heterogeneous LoRA approaches. This two-level optimization objective is formulated as:
% \begin{equation}
% \label{heter-rank-two}
% \begin{aligned}
%     \min_{\mA,\mB} \sum^K_{k=1} p_kf_k(\mA,\mB,\mA_k^*,\mB_k^*,\mW';D_k) \\
%     \text{s.t., } \mA_k^*,\mB_k^* \in \argmin_{\mA_k,\mB_k} f_k(\mA,\mB,\mA_k,\mB_k,\mW';D_k),
% \end{aligned}
% \end{equation} 
Moreover, full-size matrix aggregation and rank clustering methods %\cite{bai2024federated,fang2024automated,wang2024flora,qin2024fedinc} 
(Section~\ref{sec-aggre}) can naturally accommodate heterogeneous ranks for personalization, because full-size matrix $\Delta \mW$ retains a fixed size regardless of individual ranks and rank clustering treats rank as the fundamental unit for aggregation, ensuring the compatibility of heterogeneous rank in FedLoRA.

\subsection{Personalized LoRA}
\label{sec-personalized-lora}
Another promising approach for personalization is to introduce an additional personalized LoRA alongside the global LoRA, enhancing client-specific adaptation while preserving global knowledge sharing. This section categorizes methods by how personalized LoRA is obtained and integrated in FL.
\paragraph{Dual-LoRA.} As personalization and global learning often aim to align with different data distributions, learning an additional personalized model is proposed to tackle this challenge. Specifically, for each client $k$, the framework involves learning both a global model $\vtheta$ to capture shared knowledge across all clients and a personalized model $\vtheta_k$ tailored to the client’s unique data. This dual-objective can be formulated:
\begin{equation}
\label{dual-lora}
\begin{aligned}
    \min_{\vtheta_k} f_k(\vtheta^*,\vtheta_k;D_k) 
    \text{ s.t., } \vtheta^*  \in \argmin_\vtheta \sum^K_{k=1} p_k f_k(\vtheta;D_k),
\end{aligned}
\end{equation} 
% where $\lambda$ a trade-off parameter controlling the balance between the two objectives and $L$ is a regularization function that constrains the relationship between the global and personalized parameters.
FedDPA \cite{yang2024dual} first proposed a dual-adapter framework in which one global LoRA $\vtheta=(\mA,\mB)$ is trained and aggregated across clients to capture global knowledge, while another LoRA $\vtheta_k=(\mA_k,\mB_k)$ is locally trained for personalization without communication. These two LoRAs are dynamically combined during inference to balance global generalization and local adaptation. Similarly, FDLoRA \cite{qi2024fdlora} adopted a dual-LoRA framework but incorporated periodic synchronization between the personalized and global LoRAs every few rounds, and a novel adaptive fusion method to merge these LoRAs. 
% FedQLoRA \cite{} extended the dual-LoRA framework by addressing quantization errors in heterogeneous FL scenarios. It introduced a quantization-aware learning loss to optimize the personalized LoRA, compensating for the quantization loss and improving the performance of quantized models. 
Building on the symmetry analysis of LoRA matrices, FedSA-LoRA \cite{guo2024selective} proposed a novel approach where $\vtheta=\mA$ is for global aggregation and knowledge sharing, while $\vtheta_k=\mB_k$ is reserved for local personalization, which not only addresses the personalization but also resolves the LoRA aggregation discordance. 

\paragraph{Heterogeneous Structure.} Diverging from previous works that optimize personalized LoRA with a bi-level objective, PerFIT \cite{zhang2024personalized} employed pruning-oriented neural architecture search (NAS) to discover a personalized LoRA structure tailored to each client and transformed the global aggregated LoRA into personalized LoRA with the NAS-searched structures, eliminating the need for additional optimization objectives. Therefore, for $K$ clients with LoRA $\vtheta=(\mA,\mB)$, the objective can be formulated as follows:
\begin{equation}
\label{nas-lora}
\begin{aligned}
    \min_{\vtheta,\{\gA_k\}} \sum^K_{k=1} p_k f_k(\vtheta(\gA_k);D_k) 
   \quad \text{s.t., } R_k(\gA_k) \leq B_k,
\end{aligned}
\end{equation} 
where $\gA_k$ is the personalized architecture of client $k$, and $R_k$ and $B_k$ represent the resource consumption and budget limitation for client $k$.
More recently, researchers \cite{zhang2024personalized,mei2024fedmoe} have explored mixture-of-experts with LoRA, enabling dynamic selection and combination of expert LoRAs, with $\vtheta=\{\mA_i,\mB_i\}^N_{i=1}$ denoting the set of expert LoRAs and $\gA_k$ denoting the selected experts for client $k$, enhancing both global performance and personalization.
% to train multiple expert LoRA modules. These models aim to address heterogeneity by enabling a dynamic selection and combination of experts, with $\vtheta=\{\mA_i,\mB_i\}^N_{i=1}$ representing the set of expert LoRAs and $\gA_k$ denoting the specific experts selected for client $k$. 
% This framework allows for tailoring to specific client distributions, enhancing both global generalization and personalized adaptation in FL.

\paragraph{Hypernetwork.} Beyond previously discussed approaches, HyperFloRA \cite{lu2024hyperflora} introduced hypernetworks to generate personalized LoRA parameters for each client based on its unique representation vector. Specifically, each client $k$ is assigned an indicator vector $\vr_k$ derived from its local data $D_k$, and the server trains the hypernetwork $\psi$ to generate personalized LoRA for each client with the following objective:
\begin{equation}
\label{hypernetwork-lora}
\begin{aligned}
    \min_{\psi} \sum^K_{k=1} p_k f_k(h(\psi;\vr_k);D_k).
\end{aligned}
\end{equation} 
This enables HyperFloRA to efficiently personalize models, particularly advantageous for training-incapable clients.

\subsection{Clustering} 
\label{sec-hetero-cluster}
% \paragraph{Cluster.} 
Clustering-based approaches offer another promising solution for addressing heterogeneity in FL by grouping clients with similar data distributions or preferences. In this framework, clients are assigned to $N$ clusters corresponding to a set of LoRA $\{\vtheta_n=(\mA_n,\mB_n)\}_{n=1}^N$ for optimization:
\begin{equation}
\label{cluster-lora}
\begin{aligned}
    & \min_{\{\vtheta_n\}}  \sum_{k=1}^K\sum_{n=1}^N \alpha_{k,n} p_k f_k(\vtheta_n;D_k) \\
    \text{s.t., } \quad &\alpha_{k,n} \in \argmin_{\alpha_{k,n}}\sum_{k=1}^K\sum_{n=1}^N \alpha_{k,n} d(\vtheta_k,\vtheta_n) ,
\end{aligned}
\end{equation} 
where $\alpha_{k,n}$ is the assignment matrix ( $\alpha_{k,n}=1$ if $k\in n$ else $\alpha_{k,n}=0$), $\vtheta_n$ is the centroid of cluster $n$, and $d$ is the distance function to measure the distance between a client's parameter and the cluster centroid.
Recognizing that data from the same task shares similar distributions, FL-TAC \cite{ping2024fl} proposed a server-side clustering approach to group similar LoRA modules, enabling task-specific aggregation and improving model performance for diverse tasks. Similarly, FedLFC \cite{guo2024fedlfc} introduced a clustering strategy for LoRA modules based on different languages, effectively addressing the challenges of Multilingual FL by tailoring the aggregation process to language-specific characteristics. FedHLT \cite{guo2024fedhlt} extended FedLFC by incorporating a hierarchical language tree for Multilingual FL, where the server maintains a set of LoRA parameters for each node in the language tree and aggregates clients' LoRA based on their positions in the tree, ensuring more structured and efficient knowledge sharing across related languages.
\begin{table*}
\resizebox{\linewidth}{!}{
\small
    \centering
    \setlength\tabcolsep{1.5pt} 
    \begin{tabular}[c]{ll|ll|l}
        \toprule
        & Methods  & Advantages & Disadvantages & Reference \\
        \midrule
        \multirow{10}{*}{\rotatebox{90}{{Distributed Learning}}} & FedAVG & Simple, widely used in FL & Causes LoRA aggregation discordance & {\tiny \cite{mcmahan2017communication}} \\
        % \cmidrule[0.1pt](r){2-2} \cmidrule[0.1pt](r){3-4} \cmidrule[0.1pt](r){5-5}
        &  Single Low-Rank Matrix & Simple, efficient  & Slow convergence & {\tiny \makecell[l]{\cite{sun2024improving,nguyen2024towards},\\ \cite{chen2024robust,koo2024towards}}}\\
        % \cline{2-5}
        &  Full-size Matrix & Faster convergence, supports heterogeneous ranks & Higher computational costs & {\tiny \makecell[l]{\cite{bai2024federated,fang2024automated},\\ \cite{wang2024flora}}} \\
        % \cline{2-5}
        &  Corrective Mechanism & Improves initialization consistency & High computation overhead, additional parameters & {\tiny \cite{bian2024lora,singhal2024exact}} \\
        % \cline{2-5}
        & Rank Clustering & Fine-grained aggregation, supports heterogeneous rank & Additional hyperparameters, high computation costs & {\tiny \cite{qin2024fedinc}} \\
        % \cline{2-5}
        & Server Standard Initial & Simple to implement & Introduce weight update drift & {\tiny \cite{hu2021lora}} \\
        & Server Data-Driven Initial & Enhanced data alignment & Additional computation costs & {\tiny \cite{babakniya2023slora,yan2024federa}}\\
        & Client Avg-Initial & Balances stability and adaptability &  LoRA aggregation discordance & {\tiny \cite{bian2024lora}} \\
        & Client Re-Initial & Learn of a higher-rank space & Introduce initialization drift & {\tiny \cite{bian2024lora,park2024communication}} \\
        & Client Local-Initial & Customizes initialization & Lack global consistency & {\tiny \cite{bian2024lora}}\\
        \midrule
          \multirow{5}{*}{\rotatebox{90}{{Heterogeneity}}}  & Heterogeneous Rank & Simple, supports resource heterogeneity &  Difficult to automatically select rank & \setlength{\baselineskip}{0.4\baselineskip}{\tiny \makecell[l]{\cite{cho2024heterogeneous,chen2024rbla},\\ \cite{byun2024towards}}} \\
       & Dual-LoRA & Balances personalization and global learning & Additional parameters & {\tiny \makecell[l]{\cite{yang2024dual,qi2024fdlora}, \\ \cite{guo2024selective}}} \\
        & Heterogeneous Structure & Offers flexibility, supports resource heterogeneity & Implementation complexity increases with model size & {\tiny \makecell[l]{\cite{zhang2024personalized,zhang2024personalized},\\ \cite{mei2024fedmoe}}} \\
        & Hypernetwork & Generalized to new clients &  Difficult to optimize and train effectively & {\tiny \cite{lu2024hyperflora}}\\
         & Cluster & Good for application &  High computation costs, additional hyperparameters & {\tiny \makecell[l]{\cite{ping2024fl,guo2024fedlfc},\\ \cite{guo2024fedhlt}}}\\
         \midrule
          \multirow{3}{*}{\rotatebox{90}{{Efficiency}}} & Sparse Learning & Advanced LoRA specific techs & Slow convergence & {\tiny \makecell[l]{\cite{kuo2024federated,su2024federated},\\ \cite{liu2024fisher,zhang2024fed},\\ \cite{wu2024fedfmsl,su2025fedra}}} \\
          & Split Learning & Good for large model's application & Synchronization and load-balancing issues  & {\tiny \makecell[l]{\cite{wang2024federated,zhao2024fedsllm},\\ \cite{lin2024splitlora}}} \\
          & Compression & Compatible with existing FL frameworks & Risk performance degradation if improper optimized & {\tiny {\cite{wu2024cg,wu2024fedbiot}}}\\
        \bottomrule
    \end{tabular}}
    \caption{Comparison of different FedLoRA methods for distributed learning, heterogeneity and Efficiency.}
    \label{tab:hetero}
\end{table*}

\subsection{Discussion} 
Personalization has emerged as a promising approach to mitigate heterogeneity in FedLoRA, where each approach presents trade-offs as summarized in Table~\ref{tab:hetero}. Future research should focus on developing more LoRA-derived personalization strategies like enabling automatic rank selection, as well as enhancing the generalization ability of personalized FedLoRA across diverse client distributions to align with the versatile demands of foundation models in FL applications.
%methods like Heterogeneous Rank (Section~\ref{sec-hetero-rank}) dynamically adjust LoRA ranks based on client resources, personalized LoRA (Section~\ref{sec-personalized-lora}) learns an additional LoRA for improved adaptation, and clustering methods (Section~\ref{sec-hetero-cluster}) facilitating group-based applications. Overall, each approach presents trade-offs in addressing heterogeneity in FedLoRA, as summarized in Table~\ref{tab:hetero}.

\section{Efficiency}
Despite FedLoRA's efficiency in learning with fewer parameters, modern foundation models still pose significant storage, communication, and computational challenges. Recent research addresses these issues by developing advanced efficiency-enhancing methods, which can be categorized into three key aspects based on the technologies employed.
% To address these challenges, recent research has concentrated on developing advanced methods to enhance the efficiency and scalability of learning in FedLoRA, which can be categorized into three key aspects based on the technologies employed.

\begin{figure}
    \centering
    \includegraphics[width=\linewidth]{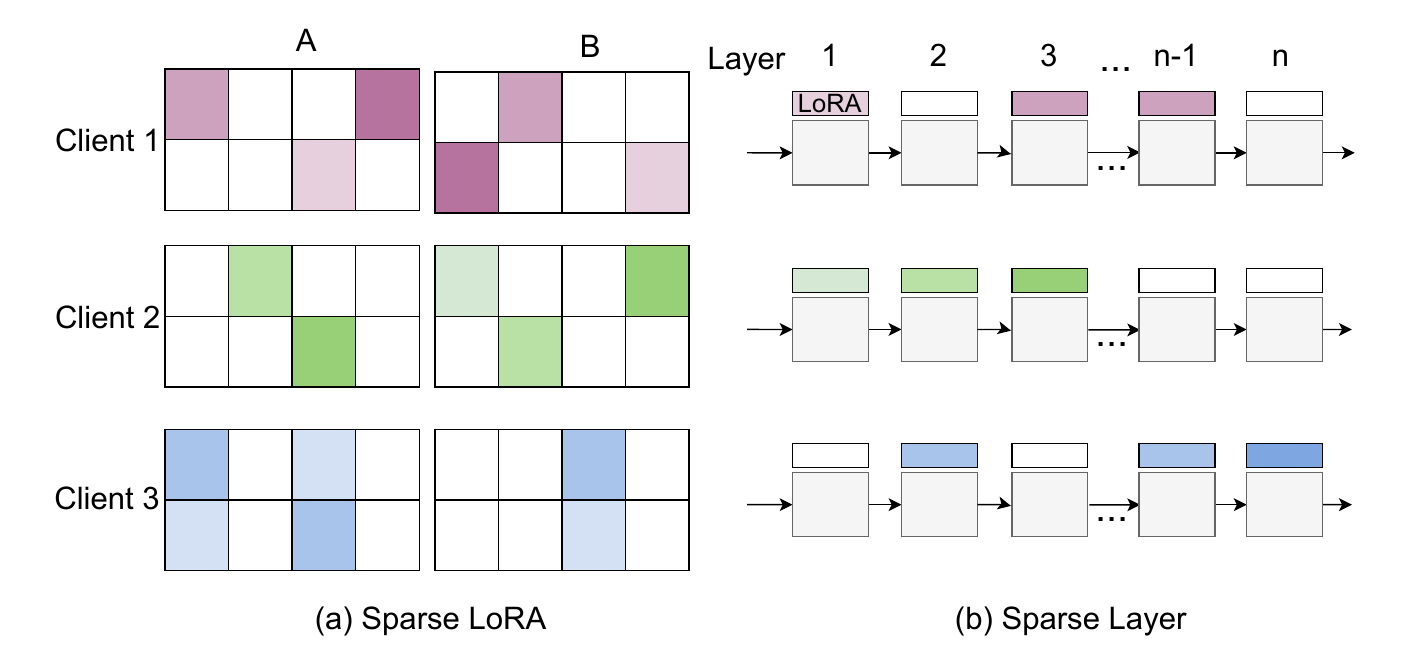}
    \vspace{-20pt}
    \caption{Two types of sparse learning frameworks in FedLoRA.}
    \label{fig:fedlora-sparse}
\end{figure}

\paragraph{Sparse Learning.}
Sparse learning is an effective approach to enhancing the efficiency of learning in FL by identifying and utilizing only the most essential parameters. One type of sparse learning focuses on applying sparsity at the LoRA parameter level, as illustrated in Figure~\ref{fig:fedlora-sparse} (a). FLASC \cite{kuo2024federated} first employed pruning methods to transmit sparsified LoRA parameters, reducing communication overhead while allowing clients to locally fine-tune the entire LoRA module for superior utility with minimal computational overhead compared to sparse tuning methods. To further balance efficiency and heterogeneity in FedLoRA, HAFL \cite{su2024federated} introduced a method that selectively updates only the most important decomposed rank-1 LoRA matrices while keeping the rest frozen, allowing clients to have heterogeneous ranks for resource-aware optimization and task-specific alignment.
Another type of sparse learning applies sparsity at the layer level, as shown in Figure~\ref{fig:fedlora-sparse} (b). To enhance both communication and computation efficiency, FibecFed \cite{liu2024fisher} and Fed-piLot \cite{zhang2024fed} proposed selecting the most important layers of LoRA for tuning and aggregation based on Fisher Information or Local-Global Information Gain Scores. Similarly, FedFMSL \cite{wu2024fedfmsl} introduced a sparsely activated LoRA layer framework, which progressively tunes specific LoRA layers over communication rounds based on training accuracy, optimizing resource usage dynamically.
To better align with clients’ resource constraints in real-world applications, FedRA \cite{su2025fedra} proposed randomly selecting layers to construct client-specific local models, tailored to their computational and storage capabilities using an allocation matrix. 

\begin{figure}
    \centering
    \includegraphics[width=\linewidth]{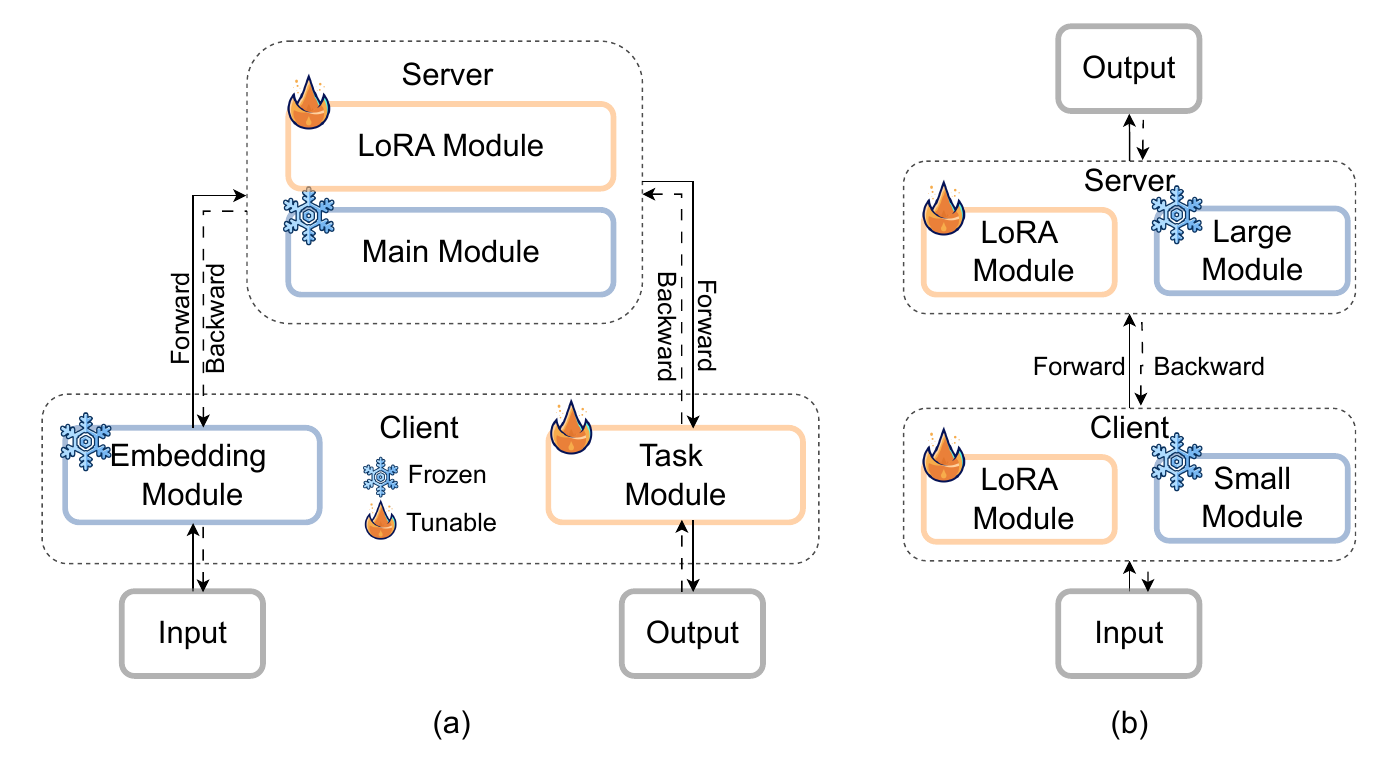}
    \vspace{-20pt}
    \caption{Two types of split learning frameworks in FedLoRA.}
    \label{fig:fedlora-split}
\end{figure}

\paragraph{Split Learning.} 
% As modern foundation models continue to grow in size and importance, it becomes essential to explore efficient methods in FL, where resource-constrained clients struggle to maintain and tune large models locally. 
As resource-constrained clients struggle to maintain and tune large models locally, the study \cite{wang2024federated} integrated a split learning framework into FedLoRA as shown in Figure~\ref{fig:fedlora-split} (a), where only the embedding and task-specific modules are retained on clients, while the main model body, comprising the majority of parameters, is hosted on the server for efficient training.
Similarly, FedsLLM \cite{zhao2024fedsllm} and SplitLoRA \cite{lin2024splitlora} also extended FedLoRA with split learning while distributing the first several layers of model to clients, placing the majority of the remaining layers to the server, and introducing a local aggregation server for efficient client-side LoRA aggregation, as shown in Figure~\ref{fig:fedlora-split} (b).

\paragraph{Compression.} %In addition to the aforementioned methods, various other approaches have been proposed to enable efficient learning in FedLoRA. 
In addition to previous methods, CG-FedLLM \cite{wu2024cg} introduced an autoencoder-based compression framework, where clients encode gradient features locally and transmit compressed representations to the server for decoding, effectively reducing communication overhead. Similarly, FedBiOT \cite{wu2024fedbiot} applied model compression techniques to obtain a compact model for tuning in FedLoRA, further leveraging knowledge distillation to align its performance with that of a fully fine-tuned model.
% integrated an autoencoder to compress communicated gradient features, allowing clients to encode gradients locally and transmit the compressed gradients to the server for decoding, significantly reducing communication overhead. Similarly, FedBiOT \cite{wu2024fedbiot} employed compression techniques to obtain a compact model for tuning in FedLoRA, while leveraging knowledge distillation to align its performance with that of the full model. %Furthermore, FwdLLM \cite{xu2024fwdllm} introduced a backpropagation-free training method for efficient learning, where clients perform perturbed inferences and use output validation for directional determination, eliminating the need for backpropagation and thereby reducing computational costs.

\paragraph{Discussion.} With the increasing scale of foundation models, recent research has focused on adapting advanced efficiency methods in FedLoRA to address computational and communication challenges, as summarized in Table~\ref{tab:hetero}. 
% Sparse learning with LoRA parameters or layers effectively reduces resource costs but may face convergence issues due to limited trainable parameters, split learning with FedLoRA enables scalable model training for application while posing synchronization and load-balancing challenges, and compression techniques adapted in FedLoRA further minimizes communication overhead but may risk performance degradation if not properly optimized. 
Despite these, FwdLLM \cite{xu2024fwdllm} introduced a backpropagation-free approach in FedLoRA for computational efficiency by replacing backpropagation with perturbed inferences and output validation. However, further research could explore more advanced methods, particularly LoRA-specific optimizations for large-scale FL scenarios, and also consider real-world applications for FedLoRA like healthcare analytics to enhance scalability and practical deployment.

% \section{Application}
% FedLoRA integrates LoRA into FL as a promising and efficient approach to address computational and communication challenges while maintaining model performance. This section highlights the applications of FedLoRA across various domains.
% For NLP applications, FedLoRA has been implemented with language models with language models to enable efficient fine-tuning, improving language understanding and generation capabilities \cite{zhang2022federated,ye2024openfedllm,jiang2024low,fan2024fedmkt,wu2024client,zhang2024towards,rao2024less,elbakary2024mira}.
% For CV applications, FedLoRA serves as a foundational framework to achieve efficient learning without sacrificing performance in tasks such as image classification and segmentation \cite{alkhunaizi2024probing,yang2024sa,asokan2024federated}.
% For multi-modal tasks, FedLoRA facilitates efficient learning across diverse modalities, enhancing the integration and alignment of information from different data types to support complex tasks such as vision-language understanding \cite{nguyen2024flora,ma2024modality}, speech-to-text translation \cite{du2024communication} and multi-sensor fusion \cite{wang2024towards}.
% For recommendation systems, FedLoRA enables the development of personalized recommendation models by efficiently learning LoRA parameters tailored to user preferences to reduce communication and computational costs, making it well-suited for large-scale recommendation scenarios \cite{zhang2024federated,nguyen2024towards}.

\section{Future Directions}
\paragraph{Theory Analysis.} 
Existing work has empirically validated the effectiveness of various FedLoRA methods, but their theoretical convergence remains underexplored. Recently, a study \cite{malinovsky2024randomized} analyzed the convergence rates of FedLoRA with different optimizers, and the other work \cite{mahla2024gradient} highlighted the potential instability of previous FedLoRA methods via convergence analysis.
While these provide initial theoretical insights, further research is needed to rigorously examine FedLoRA's convergence properties, particularly under heterogeneous settings, considering factors such as different aggregation algorithms, personalization models, and initializations. 

\paragraph{LoRA-derived Methods.} 
Recently, various enhanced LoRA methods (Section~\ref{loras}) have emerged. These advancements encourage future research to adapt them within FL to develop more sophisticated FedLoRA approaches. For example, LoRA with adaptive rank adjustment \cite{valipour2022dylora,zhang2023adalora} could be employed to dynamically adjust the LoRA rank with individual client's need for heterogeneity, while pruning enhanced LoRA \cite{dettmers2024qlora,xu2023qa} could be integrated into FedLoRA to further reduce communication and computational costs. Future research could explore the integration of these enhanced LoRA methods within FedLoRA and propose derived frameworks to further improve efficiency and performance.
% \subsection{Generalization}

\paragraph{Unified Benchmark.} As LoRA is a PEFT method for large foundation models, datasets and models used in these foundation models are highly diverse, leading to significant benchmark variations in FedLoRA studies. Unlike conventional FL, which often uses standardized datasets like ImageNet \cite{deng2009imagenet}, FedLoRA lacks a unified benchmark and metrics for fair comparison of existing approaches. Moreover, its definition of heterogeneity remains underexplored, as foundation models face broader heterogeneity beyond label distribution shifts \cite{ren2024advances}. Future research should focus on standardizing benchmarks and refining heterogeneity definitions for fair evaluation and broader applicability.

\paragraph{Various Application.}
While FedLoRA has shown promise in general machine learning tasks like image classification and language understanding, its potential in other domains remains underexplored. Recent studies \cite{zhang2024federated,nguyen2024towards} have applied FedLoRA to recommendation systems, enabling personalized models with LoRA learning for efficiency at scale. Beyond this, FedLoRA could be leveraged in other areas, like weather prediction and financ to efficiently process time series data with long history. Further exploration of FedLoRA in these domains could unlock new opportunities for efficient and secure model development.

\section{Conclusion}
As modern foundation models grow in size and complexity, FedLoRA has been proposed by integrating LoRA into FL for efficient learning. This paper provides a comprehensive survey of FedLoRA, primarily focusing on distributed learning, heterogeneity and efficiency challenges, further categorizing existing research into granular subcategories based on applied technologies. For each subcategory, we introduce underlying methodologies with detailed mathematical formulations and Figures, comparing their advantages and limitations. Finally, we discuss promising future directions for FedLoRA to guide further research and development.

% \subsection{Footnotes}

% Place footnotes at the bottom of the page in a 9-point font.  Refer to
% them with superscript numbers.\footnote{This is how your footnotes
%     should appear.} Separate them from the text by a short
% line.\footnote{Note the line separating these footnotes from the
%     text.} Avoid footnotes as much as possible; they interrupt the flow of
% the text.

% \begin{algorithm}[tb]
%     \caption{Example algorithm}
%     \label{alg:algorithm}
%     \textbf{Input}: Your algorithm's input\\
%     \textbf{Parameter}: Optional list of parameters\\
%     \textbf{Output}: Your algorithm's output
%     \begin{algorithmic}[1] %[1] enables line numbers
%         \STATE Let $t=0$.
%         \WHILE{condition}
%         \STATE Do some action.
%         \IF {conditional}
%         \STATE Perform task A.
%         \ELSE
%         \STATE Perform task B.
%         \ENDIF
%         \ENDWHILE
%         \STATE \textbf{return} solution
%     \end{algorithmic}
% \end{algorithm}

% \appendix

% \section*{Ethical Statement}

% \section*{Acknowledgments}

\scriptsize
\bibliographystyle{named}
\bibliography{ijcai25}

\end{document}